# Kinematics of Spherical Robots Rolling Over 3D Terrains

Saeed Moazami[1], Hassan Zargarzadeh[1], and Srinivas Palanki[2],

[1]S. Moazami and H. Zargarzadeh are with the Phillip M. Drayer Department of Electrical Engineering at Lamar University, Beaumont, TX, USA. (e-mails: smoazami@lamar.edu, h.zargar@lamar.edu).

[2]S. Palanki is with the Dan F. Smith Department of Chemical Engineering at Lamar University, Beaumont, TX, USA. (e-mail: spalanki@lamar.edu).

Although the kinematics and dynamics of spherical robots (SRs) on flat horizontal and inclined 2D surfaces are thoroughly investigated, their rolling behavior on generic 3D terrains has remained unexplored. This paper derives the kinematics equations of the most common SR configurations rolling over 3D surfaces. First, the kinematics equations for a geometrical sphere rolling over a 3D surface are derived along with the characterization of the modeling method. Next, a brief review of current mechanical configurations of SRs is presented as well as a novel classification for SRs based on their kinematics. Then, considering the mechanical constraints of each category, the kinematics equations for each group of SRs are derived. Afterwards, a path tracking method is utilized for a desired 3D trajectory. Finally, simulations are carried out to validate the developed models and the effectiveness of the proposed control scheme.

## 1. Introduction

Spherical robots are a class of mobile robots that are generally recognized by their ball-shaped shell and internal driving components that provide torques required for their rolling motion. From the ball-shaped exterior, SRs inherit multiple advantages over other types of mobile robots, in which skidding, tipping over, falling, or friction with the surface makes them vulnerable or inefficient [1, 2]. Despite all the unique features, the complicated nonlinear behavior of SRs has remained as a hurdle to fully comprehend their dynamics, motion kinematics, and unveil their maneuverability capabilities.

Although the earliest efforts to analytically capture the kinematics and dynamics of rolling geometrical balls on mathematical surfaces back to more than a century ago e.g. in E. Routh and S.A. Chaplygin's works, this topic is still an open discussion that is being investigated in more recent papers such as [3-5]. Numerous researches have been conducted on the mathematical modeling of kinematics and dynamics of SRs with a variety of mechanical configurations [6, 7]. Among them, a widespread assumption is that SRs roll over an ideally flat horizontal plane. In a number of available works, where SRs are studied on non-horizontal surfaces, simplifying assumptions are made. For instance, in [2, 8], two SR designs are investigated that climb obstacles, assuming the condition to be static. From a different view, rolling of SRs are studied where the desired path is assumed to be a straight line with constant slope or a single step obstacle [9-11] and a 2D curved path with variable-slope [12] respectively. In [13], authors have investigated dynamics of Martian tumbleweed rovers while this special type of SR rolls in its heading direction and the turning action is not considered for them. In fact, while several researches have been done on 3D kinematics of other types of mobile robots such as legged [14] and wheeled robots [15, 16], to the best of the authors' knowledge, the general problem of kinematics of SRs rolling on 3D terrains has not been investigated in the literature. The motivation to address this problem is that, while many applications of the SRs are on flat surfaces such as indoor [17], and paved roads [18], for a variety of applications such as agriculture [19], surveillance [20], environmental monitoring [21], and even planetary explorations [22], they would get exposed to uneven terrains.

In this work, prior to deriving the kinematics of SRs on 3D terrains, a general method for modeling a geometrical sphere rolling over a mathematically known 3D surface is developed. Then, the derived equations are expanded in order to be applied to SRs considering their specifications. Concretely, a variety of mechanisms are utilized in SRs to provide the required propelling torques and forces for their rolling action. Each configuration imposes its own kinematical constraints on the SRs' rolling motion. Therefore, to study the kinematics of SRs it is essential to classify current and feasible designs of SRs accordingly.

There are a few SR classifications available in the literature. In a survey [6], SRs are classified based on their mechanical driving principles as: 1) Barycenter offset (BCO), 2) Conservation of angular momentum (COAM), and 3) Shell transformation (OST). In another review, SRs are classified based on their mechanical configurations [23], e.g., single wheel, hamster wheel, pendulum driven, gimbal mechanism, single ball, mass movement, and a set of designs that use flywheels. Again, each class of the above robots can have different kinematics. Therefore, to investigate the rolling behavior of SRs based on the constraints imposed by their kinematical configurations, presenting a new, although brief, classification seems inevitable.

The classification proposed in this work, divides SRs into two major categories: 1) Continuous Rolling Spherical Robots (CR-SR), with triple axes rolling (3R-SR), dual axes rolling (2R-SR), and rolling and turning (RT-SR) spherical robots as sub-categories, and 2) Rolling and Steering Spherical Robots (RS-SR). Based on the constraints of each different category, the kinematics equations for SRs rolling over 3D terrains are derived.

Afterwards, a modified pure pursuit method [24] is utilized for the path tracking control problem of the proposed models. The path tracking method indicates the required kinematic actuation states that should be provided in each SR categories to track a desired 3D trajectory. Finally, simulations are carried out to verify the kinematics model's accuracy and its controller's tracking efficiency in different models along with comparison between their behavior.

The remainder of this paper is organized as follows: the next section is devoted to the model description. The third section elaborates the problem of kinematics modeling of a sphere rolling over a 3D surface. In section four, the objectives of the paper are addressed, including presenting a new classification for SRs by their kinematics properties and then deriving kinematics equations of SRs based on specifications of each selected SR category. In section five, a 3D path tracking kinematic control scheme is presented. The simulation results in MATLAB are represented in section six, followed by the conclusion and future work.

## 2. Model Description

Consider a geometrical sphere rolling over a 3D surface $\mathcal{S}$ as shown in Fig. 1. To derive the kinematics equations of the rolling motion, the first step is to define the required reference frames, as $\{\mathcal{F}_W\}=\{O_W, X_{\bar{W}}, Y_{\bar{W}}, Z_{\bar{W}}\}$, $\{\mathcal{F}_{Tr}\}=\{O_{Tr}, x_{\bar{Tr}}, y_{\bar{Tr}}, z_{\bar{Tr}}\}$, $\{\mathcal{F}_{T_0}\}=\{O_{T_0}, x_{\bar{T}_0}, y_{\bar{T}_0}, z_{\bar{T}_0}\}$, $\{\mathcal{F}_T\}=\{O_T, x_{\bar{T}}, y_{\bar{T}}, z_{\bar{T}}\}$, and $\{\mathcal{F}_L\}=\{O_L, x_{\bar{L}}, y_{\bar{L}}, z_{\bar{L}}\}$, defined as world, translated, surface-side tangent, robot-side tangent, and local reference frames respectively. In each reference frame set, the first element shows its origin and the rest are the reference frame's axes. Accordingly, as it is illustrated in Fig. 1, $\{\hat{I}_W, \hat{J}_W, \hat{K}_W\}$, $\{\hat{i}_{Tr}, \hat{j}_{Tr}, \hat{k}_{Tr}\}$, $\{\hat{i}_{T_0}, \hat{j}_{T_0}, \hat{k}_{T_0}\}$, $\{\hat{i}_T, \hat{j}_T, \hat{k}_T\}$, and $\{\hat{h}, \hat{l}, \hat{n}\}$, represent unit vectors of $\{\mathcal{F}_W\}$, $\{\mathcal{F}_{Tr}\}$, $\{\mathcal{F}_{T_0}\}$, $\{\mathcal{F}_T\}$, and $\{\mathcal{F}_L\}$ respectively.

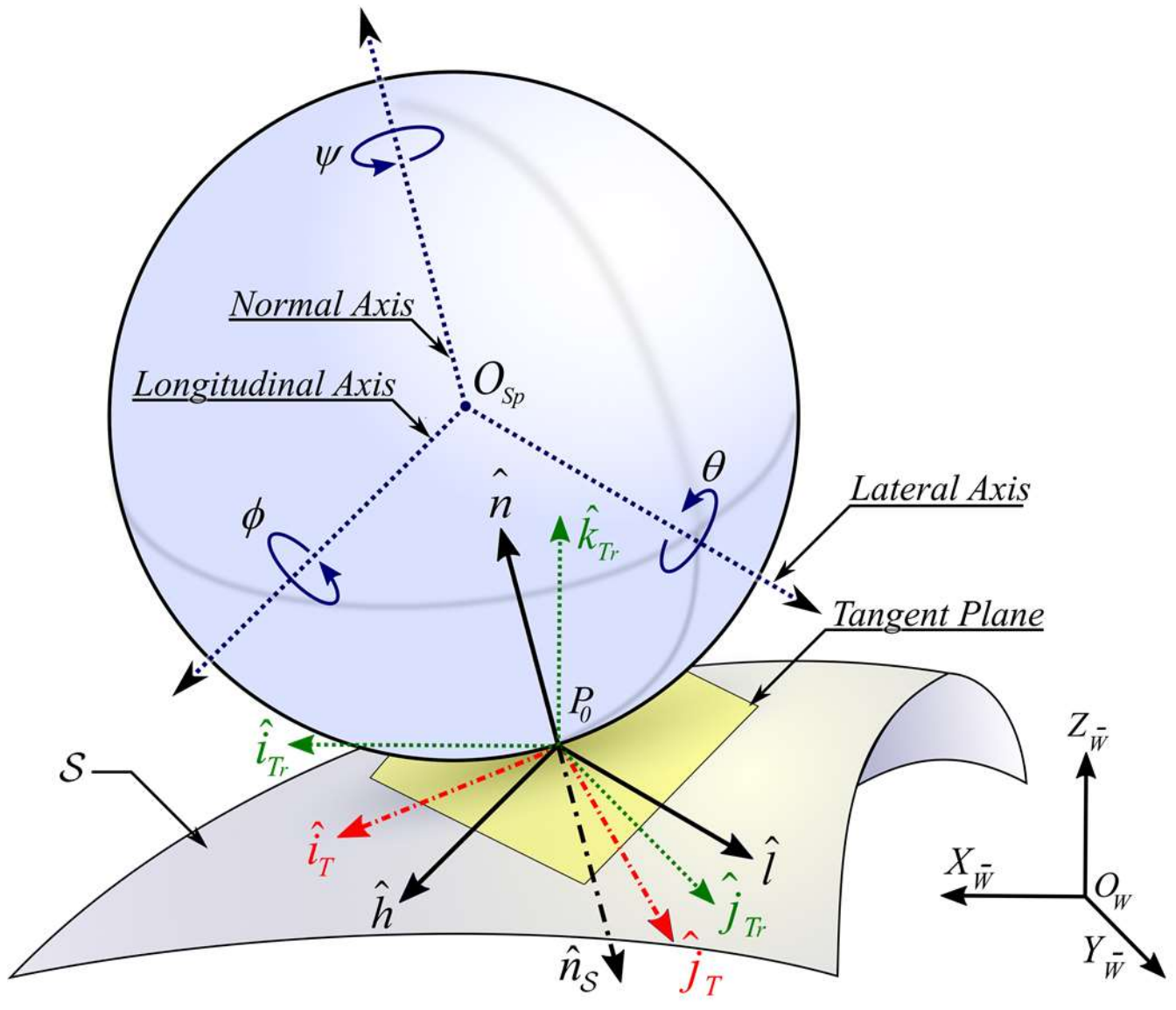

Fig. 1. Schematic diagram of reference frames and kinematic parameters that are used to derive kinematics equation of rolling sphere on a 3D surface $\mathcal{S}$.

In the sequel, vector quantities are defined in $\{\mathcal{F}_W\}$, if otherwise, their reference frames are specified in left superscript, e.g. ${}^{\mathcal{F}_A}\vec{V}$. Vectors are represented by an overhead right arrow while unit vectors are accented by a caret shape hat symbol. Homogeneous transformations e.g. from arbitrary reference frames $\{\mathcal{F}_A\}$ to $\{\mathcal{F}_B\}$, consisting of a rotation matrix of $\mathcal{R}_{AB(3\times3)}$ and a translation vector $\mathcal{M}_{AB(3\times1)}$, are defined in the form of [25]:

$$\mathcal{T}_{AB} = \left[\begin{array}{c|c} \mathcal{R}_{AB} & \mathcal{M}_{AB} \\ \hline 0_{1\times3} & 1 \end{array}\right]. \tag{1}$$

Through utilizing transformation matrices, a fourth element of $0$ is added to velocity vectors as translation does not apply to them. A fourth element of $1$ in position vectors guarantees both rotation and translation.

In Fig. 1, the position of the contact point of the sphere with the 3D surface $\mathcal{S}$ is represented as $P_0$:

$$P_0 = {}^{\mathcal{F}_W}P_0 = [x_0(t), y_0(t), z_0(t)]^T, \tag{2}$$

that instantaneously indicates the position coordinates of $O_{Tr}$, $O_{T_0}$, $O_T$, and $O_L$. The imaginary plane tangent to $\mathcal{S}$ passing through $P_0$ is called tangent plane denoted as $T_S$. As it is shown in Fig. 1, normal axis of the sphere is defined to be perpendicular to $T_S$ coinciding with the center of the sphere, $O_{Sp}$, while lateral and longitudinal axes are defined to be parallel to $T_S$, i.e., they do not rotate through rotation of the sphere about axes parallel to $T_S$, but they rotate accordingly when the sphere rotates about the normal axis. $\{\mathcal{F}_{Tr}\}$ is resulted from translation of $\{\mathcal{F}_W\}$ to $P_0$ such that $\vec{x}_{Tr}$, $\vec{y}_{Tr}$ and $\vec{z}_{Tr}$ remain parallel to $\vec{X}_W$, $\vec{Y}_W$ and $\vec{Z}_W$. Obviously, axes of $\{\mathcal{F}_{Tr}\}$ do not necessarily lie on $T_S$. In $\{\mathcal{F}_{T_0}\}$, $\vec{z}_{T_0}$ is along the normal vector of $\mathcal{S}$, $\vec{n}_S$. $\vec{x}_{T_0}$ is locally tangent to the surface and therefore lies on $T_S$. $\vec{y}_{T_0}$ is mutually perpendicular to $\vec{x}_{T_0}$ and $\vec{z}_{T_0}$ following the right-hand rule. Further, as shown in Fig. 1, $\{\mathcal{F}_T\}$ is resulted from rotation of $\{\mathcal{F}_{T_0}\}$ about $\vec{x}_{T_0}$ by 180 degrees. Consequently, $\vec{x}_T$ coincides with $\vec{x}_{T_0}$. $\vec{y}_T$ and $\vec{z}_T$ are in opposite direction to $\vec{y}_{T_0}$ and $\vec{z}_{T_0}$ with respect to $O_{T_0}$. In $\{\mathcal{F}_L\}$, $\vec{x}_L$ and $\vec{y}_L$ are also tangent to $T_S$. $\vec{x}_L$ is along the sphere's longitudinal direction. $\vec{z}_L$ is normal to $T_S$ coinciding with $\vec{z}_T$ and normal axis of the sphere, and $\vec{y}_L$ is mutually perpendicular to $\vec{x}_L$ and $\vec{z}_L$ following the right-hand rule. One can note that $\vec{y}_L$ can be considered as the projection of the sphere's lateral axis on $T_S$.

The next step is to define rotation angles of the sphere. According to Fig. 1, $\theta$ is the sphere's rolling angle, i.e., the rotation angle of the sphere about its lateral axis. $\phi$ is the sphere's tilting angle, i.e., the rotation angle of the sphere about its longitudinal axis, and $\psi$ is the sphere's turning angle about the robot-side normal axis, $\hat{n}$. It can be noted that $\{\mathcal{F}_L\}$ can be considered as $\{\mathcal{F}_T\}$ rotated through an angle of $\psi$ about $\vec{z}_T$. Finally, $\mathcal{S}$ is a 3D surface defined in $\{\mathcal{F}_W\}$ as $\mathcal{S}$: $F(x,y,z)=0$, in the form of:

$$f(x,y) - z = 0, \tag{3}$$

implying that for any $x$ and $y$, the elevation of the surface $\mathcal{S}$ along the vertical axis of $\vec{Z}_W$ is $z = f(x,y)$.

*Assumptions*: The mathematical model is derived based on the following assumptions:

1) *The sphere rolls over a 3D surface without slipping, skidding or falling, i.e. it does not lose traction with the surface.*

2) *The surface $\mathcal{S}$ is smooth, i.e. $z = f(x,y)$ is a continuous function of class $C^1$ or higher [26].*

3) *The sphere is rigid, and the radius of curvature of the surface is never smaller than the radius of the sphere. Therefore, the sphere always remains in a single point contact with the surface.*

## 3. Kinematics Modeling of Spheres on 3D Surfaces

The next phase to derive the kinematics equations of the sphere rolling over $\mathcal{S}$ is to calculate the utilized reference frames' unit vectors analytically along with deriving transformation matrices between reference frames. In this process, first, we will determine the vector quantities in $\{\mathcal{F}_L\}$, and then by sequentially utilizing transformation matrices, we can calculate the desired quantities in $\{\mathcal{F}_W\}$.

As mentioned in the previous section, $\{\mathcal{F}_{Tr}\}$ is parallel to $\{\mathcal{F}_W\}$ and $O_{Tr}$ coincides with $P_0$, therefore, $\mathcal{T}_{TrW}$ performs no rotation, i.e. $\mathcal{R}_{TrW} = \mathbf{I}_3$ and a translation of $\mathcal{M}_{TrW} = P_0$. Thus, $\mathcal{T}_{TrW}$ can be written as follows:

$$\mathcal{T}_{TrW} = \begin{bmatrix} \mathbf{I}_3 & P_0 \\ 0 & 1 \end{bmatrix}, \tag{4}$$

where, $\mathbf{I}_{3\times 3}$ is the identity matrix. From (3), the normal vector of surface $\mathcal{S}$ at position $P_0$ is given by:

$$\vec{n}_{\mathcal{S}} = \nabla F = \left[ f_x, f_y, -1 \right]^T, \tag{5}$$

where, $f_x = \partial f(x,y)/\partial x$, and $f_y = \partial f(x,y)/\partial y$ are partial derivative functions. Now, as $\vec{n}_{\mathcal{S}}$ coincides with $z_{T_0}^-$, its unit vector can be written as:

$$\hat{k}_{T_0} = \hat{n}_{\mathcal{S}} = \vec{n}_{\mathcal{S}} / \| \vec{n}_{\mathcal{S}} \|. \tag{6}$$

Defining $s_n$ as the following:

$$s_n = \| \vec{n}_{\mathcal{S}} \|^{-1} = \left( 1 + f_x^{\,2} + f_y^{\,2} \right)^{-\frac{1}{2}}, \tag{7}$$

we can write:

$$\hat{n}_{\mathcal{S}} = s_n \vec{n}_{\mathcal{S}}. \tag{8}$$

Seemingly, the direction of $\hat{n}_{\mathcal{S}}$ is perpendicular to the surface and always downward, i.e., $Z_{\hat{n}_{\mathcal{S}}} = -1$. As robot-site normal, $\hat{n}$, is in the opposite direction to the surface-side normal and based on assumption (1), there is no relative velocity between $\{\mathcal{F}_{T_0}\}$ and $\{\mathcal{F}_T\}$, therefore, we can write:

$$\mathcal{T}_{TT_0} = \begin{bmatrix} 1 & 0 & 0 & 0 \\ 0 & -1 & 0 & 0 \\ 0 & 0 & -1 & 0 \\ 0 & 0 & 0 & 1 \end{bmatrix}. \tag{9}$$

In fact, $\hat{n}$ and $\hat{k}_T$ can be calculated by rotating $\vec{n}_{\mathcal{S}}$ by 180 degrees about $\hat{i}_{T_0}$, therefore:

$$\hat{n} = \hat{k}_T = -\hat{n}_{\mathcal{S}} = s_n[-f_x, -f_y, 1]^T . \tag{10}$$

Next, we calculate the unit vectors of $\{\mathcal{F}_T\}$ axes. Let us consider a sphere that rolls over $\mathcal{S}$ in a way that $\psi = 0$, that means only $\theta$ and $\phi$ contribute in the rolling action. Due to this rolling, the direction of sphere's longitudinal and lateral axes changes accordingly to match the geometry of $\mathcal{S}$. Based on the Euler's rotation theorem, any Cartesian coordinate system in 3D space with a common origin are related by a rotation about a unique axis called Euler axis denoted as $\hat{e}$. Intuitively, this rotation maps axes of $\{\mathcal{F}_T\}$ to $\{\mathcal{F}_{Tr}\}$. We know that the rotation maps $\hat{n}$ to $\hat{k}_{Tr}$ and it is assumed that $\psi = 0$, therefore, as it is depicted in Fig. 2, $\hat{e}$ should lie in the tangent plane and be perpendicular to both $\hat{n}$ and $\hat{k}_{Tr}$, leading to:

$$\vec{e} = \hat{k}_{Tr} \times \hat{n} . \tag{11}$$

Considering $\hat{k}_{Tr} = [0,0,1]^T$ and $\hat{n}$ from (10), $\vec{e}$ is calculated as:

$$\vec{e} = s_n[f_y, -f_x, 0]^T . \tag{12}$$

Consequently:

$$\hat{e} = \vec{e} / \|\vec{e}\| = s[f_y, -f_x, 0]^T \text{ with } s = \|\vec{e}\|^{-1} = 1 \Big/ \sqrt{\left(f_x^{\,2} + f_y^{\,2}\right)} . \tag{13}$$

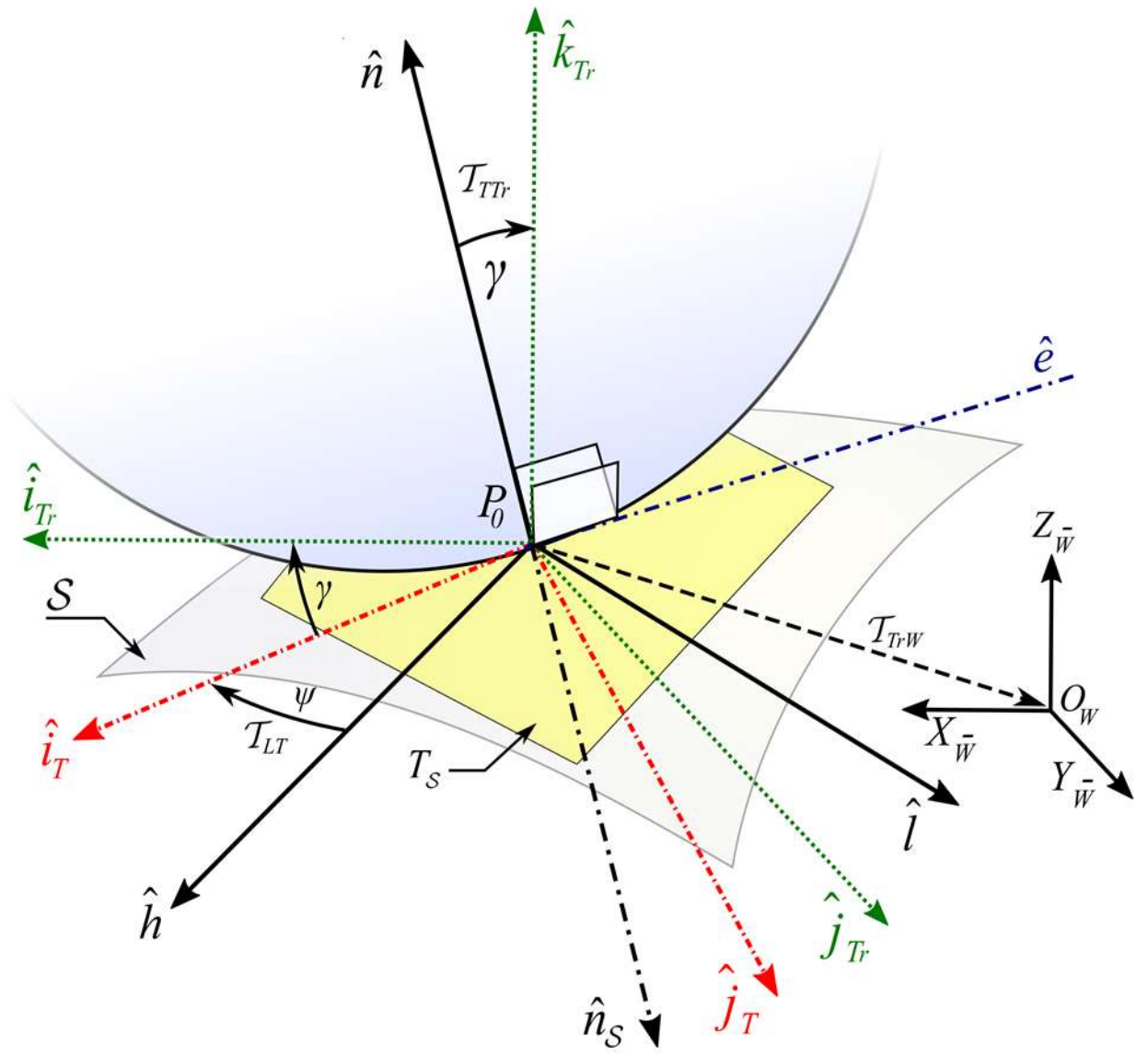

Fig. 2. Schematic diagram of reference frames, Euler axis $\hat{e}$, and transformations between reference frames.

Moreover, the angle of rotation can be calculated as:

$$\gamma = atan2(\|\hat{k}_{Tr} \times \hat{n}\|, \hat{k}_{Tr} \bullet \hat{n}) . \tag{14}$$

where *atan2* is four-quadrant inverse tangent, where *atan2* is four-quadrant inverse tangent,

$$\|\hat{k}_{Tr} \times \hat{n}\| = \sin(\gamma) = s_n s^{-1} , \text{ and } \hat{k}_{Tr} \bullet \hat{n} = \cos(\gamma) = s_n . \tag{15}$$

Therefore, to avoid singularity, from (14) and (15), $\gamma$ is written in the following conditional statement form:

$$\gamma=\begin{cases} atan2\left(s^{-1},1\right) & \text{if} \quad \hat{k}_{Tr}\neq\hat{n} \\ 0 & \text{if} \quad \hat{k}_{Tr}=\hat{n} \end{cases}. \tag{16}$$

Now, to calculate $\hat{i}_T$, we should apply the same rotation to $\hat{i}_{Tr}=[1,0,0]^T$. There are several algorithms to transform a vector-in space, given a rotation axis and an angle of rotation such as the Rodrigues' rotation method that is explained in the following.

According to Rodrigues' rotation formula [27], as $\hat{i}_{Tr}$ is a vector in $\mathbb{R}^3$ and $\hat{e}=[e_x,e_y,0]^T$ is a unit vector representing the axis of rotation about which $\hat{i}_T$ rotates by an angle of $\gamma$ resulting in $\hat{i}_{Tr}$, it can be shown that:

$$\hat{i}_T=\hat{i}_{Tr}\cos(\gamma)+\hat{e}\times\hat{i}_{Tr}\sin(\gamma)+\hat{e}\left(\hat{e}\bullet\hat{i}_{Tr}\right)\left(1-\cos(\gamma)\right). \tag{17}$$

Equation (17) can be written in matrix form to calculate $\mathcal{T}_{TTr}$ using Rodrigues' rotation formula. Let $E_\times$ denote cross-product matrix for $\hat{e}$ , i.e., a matrix when multiplied from the left with a vector, gives the cross product. It can be shown that:

$$E_\times=\begin{bmatrix} 0 & -e_z & e_y \\ e_z & 0 & -e_x \\ -e_y & e_x & 0 \end{bmatrix}=\begin{bmatrix} 0 & 0 & -sf_x \\ 0 & 0 & -sf_y \\ sf_x & sf_y & 0 \end{bmatrix}. \tag{18}$$

Now, it can be shown that rotation matrix $\mathcal{R}_{TTr}$ can be written in the following form:

$$\mathcal{R}_{TTr}=\mathbf{I}_3+\sin(\gamma)E_\times+\left(1-\cos(\gamma)\right)E_\times{}^2, \tag{19}$$

that, using (18), can be written as the following:

$$\mathcal{R}_{TTr}=\begin{bmatrix} 1-s^2f_x{}^2\left(1-C\gamma\right) & -s^2f_xf_y\left(1-C\gamma\right) & -sf_xS\gamma \\ -s^2f_xf_y\left(1-C\gamma\right) & 1-s^2f_y{}^2\left(1-C\gamma\right) & -sf_yS\gamma \\ sf_xS\gamma & sf_yS\gamma & 1-s^2\left(f_x{}^2+f_y{}^2\right)\left(1-C\gamma\right) \end{bmatrix}. \tag{20}$$

where, for the sake of brevity, $\sin(\bullet)$ and $cos(\bullet)$ are abbreviated to $S_\bullet$ and $C_\bullet$ respectively. Considering $S\gamma=s_ns^{-1}$ from (15), we can write:

$$sf_xS\gamma=s_nf_x \text{ , and } sf_yS\gamma=s_nf_y. \tag{21}$$

Also, form (13) and (15) it can be concluded that:

$$1-s^2\left(f_x{}^2+f_y{}^2\right)\left(1-C\gamma\right)=s_n. \tag{22}$$

As $\mathcal{M}_{TTr}=0_{(3\times1)}$, and by substituting (21) and (22) in (20), we can write:

$$\mathcal{T}_{T_0Tr}=\begin{bmatrix} \mathcal{R}_{TTr} & 0 \\ 0 & 1 \end{bmatrix}=\begin{bmatrix} 1-s^2f_x{}^2\left(1-s_n\right) & -s^2f_xf_y\left(1-s_n\right) & -s_nf_x & 0 \\ -s^2f_xf_y\left(1-s_n\right) & 1-s^2f_y{}^2\left(1-s_n\right) & -s_nf_y & 0 \\ s_nf_x & s_nf_y & s_n & 0 \\ 0 & 0 & 0 & 1 \end{bmatrix}. \tag{23}$$

Alternatively, to calculate $\mathcal{T}_{TTr}$, rotation quaternions can be utilized which is explained in details in appendix A. Depending on the application $\mathcal{R}_{TTr}$ can be calculated from (19), (23), or through quaternions. To calculate $\hat{j}_T$, we can write:

$$\hat{j}_T = \hat{n} \times \hat{i}_T . \tag{24}$$

Next, as $\{\mathcal{F}_L\}$ is resulted from the rotation of $\{\mathcal{F}_T\}$ through $\psi$ about $z_T^-$, then, $\mathcal{M}_{LT} = 0_{(3\times1)}$ and $\mathcal{T}_{LT}$ can be written as follows:

$$\mathcal{T}_{LT} = \begin{bmatrix} C\psi & S\psi & 0 & 0 \\ -S\psi & C\psi & 0 & 0 \\ 0 & 0 & 1 & 0 \\ 0 & 0 & 0 & 1 \end{bmatrix}. \tag{25}$$

Finally, from (4) , (23), and (25) it can be concluded that:

$$\mathcal{T}_{LW} = \mathcal{T}_{TrW}\mathcal{T}_{TTr}\mathcal{T}_{LT} = \begin{bmatrix} a_{11} & a_{12} & a_{13} & a_{14} \\ a_{21} & a_{22} & a_{23} & a_{24} \\ a_{31} & a_{32} & a_{33} & a_{34} \\ a_{41} & a_{42} & a_{43} & a_{44} \end{bmatrix}, \tag{26}$$

where the elements of $\mathcal{T}_{LW}$ are given as:

$$\begin{aligned}
&a_{11} = \left(1 - s^2 f_x^2 (1-s_n)\right)C\psi + s^2 f_x f_y (1-s_n) S\psi \\
&a_{12} = \left(1 - s^2 f_x^2 (1-s_n)\right)S\psi - s^2 f_x f_y (1-s_n) C\psi \\
&a_{13} = -s_n f_x \text{ , } a_{14} = x_0 \\
&a_{21} = -s^2 f_x f_y (1-s_n) C\psi - \left(1 - s^2 f_y^2 (1-s_n)\right)S\psi \\
&a_{22} = -s^2 f_x f_y (1-s_n) S\psi + \left(1 - s^2 f_y^2 (1-s_n)\right)C\psi \\
&a_{23} = -s_n f_y \text{, } a_{24} = y_0 \text{,} \\
&a_{31} = s_n f_x C\psi - s_n f_y S\psi \text{ , } a_{32} = s_n f_x S\psi + s_n f_y C\psi \text{ , } a_{33} = s_n \text{ , } a_{34} = z_0 \\
&a_{41} = a_{42} = a_{43} = 0 \text{, } a_{44} = 1 .
\end{aligned} \tag{27}$$

Now, consider the sphere that instantaneously rolls over the tangent plane and let ${}^{\mathcal{F}_L}\vec{V} = [u, v, w, 0]^T$ denote the velocity of the sphere in $\{\mathcal{F}_L\}$, then, we can write:

$${}^{\mathcal{F}_W}\vec{V} = \mathcal{T}_{LW} \, {}^{\mathcal{F}_L}\vec{V} , \tag{28}$$

where, ${}^{\mathcal{F}_W}\vec{V} = [\dot{x}, \dot{y}, \dot{z}, 0]^T$ . Consequently, the position of the sphere can be calculated by integrating (28) over time. To calculate ${}^{\mathcal{F}_L}\vec{V}$ , it can be written as a set of functions of kinematics parameters such that:

$$u = g_1(\theta, \phi, \psi, \dot{\theta}, \dot{\phi}, \dot{\psi}) \text{ , } v = g_2(\theta, \phi, \psi, \dot{\theta}, \dot{\phi}, \dot{\psi}) \text{, and } w = 0 . \tag{29}$$

Obviously, $g_1$ and $g_2$ functions vary depending on the configuration of the SR. Transforming ${}^{\mathcal{F}_L}\vec{V}$ to $\{\mathcal{F}_W\}$ by plugging (29) into (28) and using terms presented in (27) we have:

$$[\dot{x} \;\; \dot{y} \;\; \dot{z} \;\; 0]^T = \mathcal{T}_{LW} [g_1 \;\; g_2 \;\; 0 \;\; 0]^T . \tag{30}$$

Resulting in:

$$\dot{x} = a_{11} g_1 + a_{12} g_2 \text{ , } \dot{y} = a_{21} g_1 + a_{22} g_2 \text{ , and } \dot{z} = a_{31} g_1 + a_{32} g_2 . \tag{31}$$

The next section, first, presents a classification based on kinematics specifications of SRs, then, provides calculation steps of ${}^{\mathcal{F}_L}\vec{V}$ for each class of SR.

## 4. Kinematics Modeling of Spherical Robots Rolling Over 3D Surfaces

This section is dedicated to deriving the kinematics of SRs rolling over 3D surfaces. First, we present a brief classification in which the most popular configurations of SRs are divided into two main categories and a number of subcategories. Secondly, by applying constraints of each category (or type), the kinematics of SRs on 3D surfaces is derived.

### *A. Continuous Rolling Spherical Robots (CR-SR):*

CR-SRs are capable of performing rolling maneuvers continuously, i.e., when the robot rolls in any direction there is no limitation for its rotation angle. However, based on their mechanical configurations, these robots can roll in different directions based on which this category can be divided into three subcategories explained in the following:

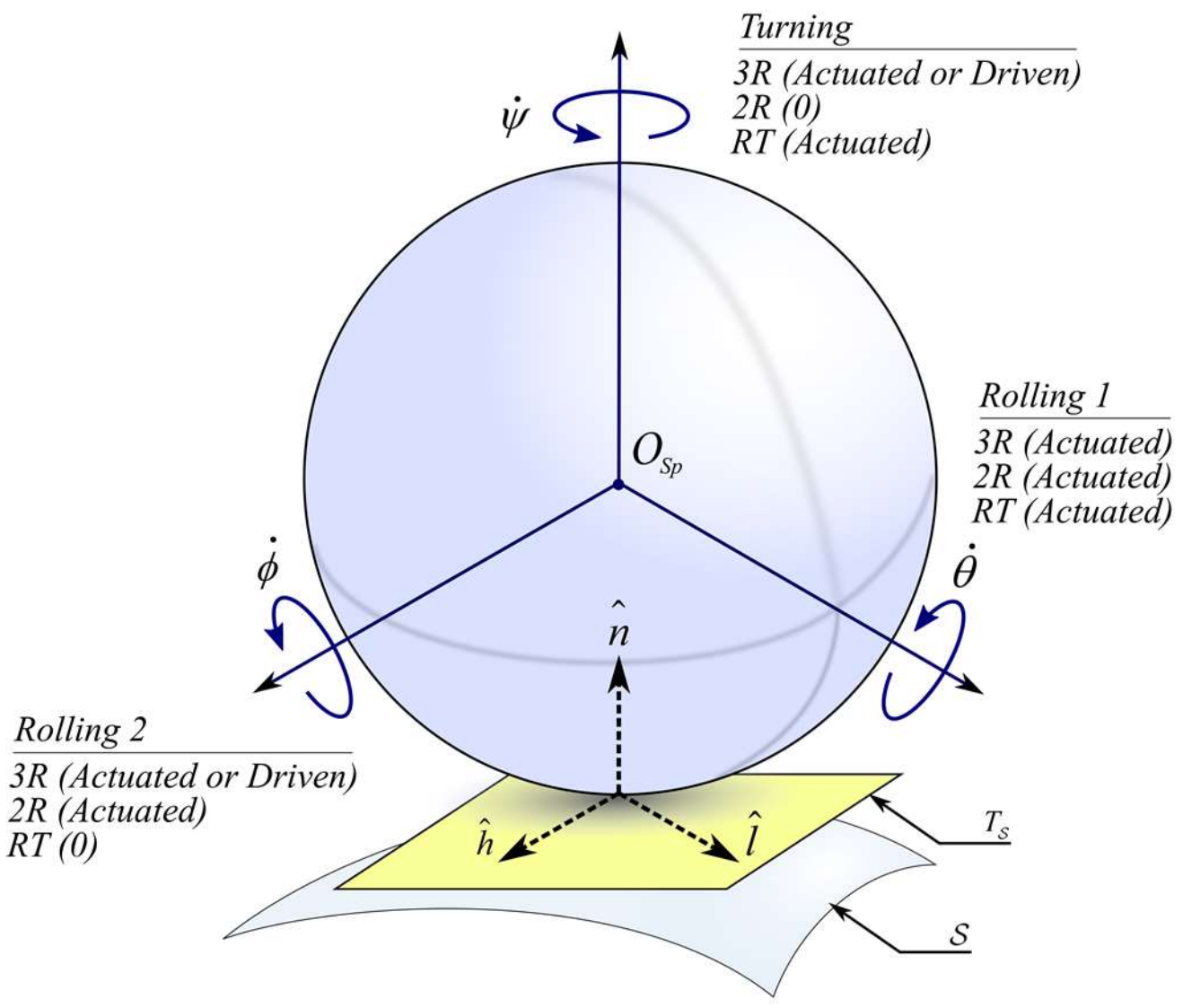

Fig. 3. Schematic diagram of rotation directions of continuous rolling spherical robots.

#### 1) *Triple-Axis Rolling Spherical Robot (3R-SR).*

This subcategory is the most general type of CR-SR. As illustrated in Fig. 3, in 3R-SRs the internal components are mechanically capable of providing torques required for the robot to rotate about all of its three main axes, namely angular velocities of $\dot{\theta}$, $\dot{\phi}$, and $\dot{\psi}$. Therefore, 3R-SRs are omni-directional. Several mechanical configurations are available in the literature that rely on this actuation method [28-30].

Fig. 4 illustrates differential rolling lengths for a 3R-SR in an infinitesimally small time $dt$ along $\hat{h}$ and $\hat{l}$ directions. Based on assumption (1) and using equal-arc-length rule for rolling without slipping [31], we have:

$$dx_L = Rd\theta\,,\ dy_L = -Rd\phi\,, \tag{32}$$

where $R$ denotes the sphere's radius. Integrating (32) over time, ${}^{\mathcal{F}_L}\vec{V}$ can be calculated as the following:

$${}^{\mathcal{F}_L}\vec{V} = [R\dot{\theta}, -R\dot{\phi}, 0]^T\,. \tag{33}$$

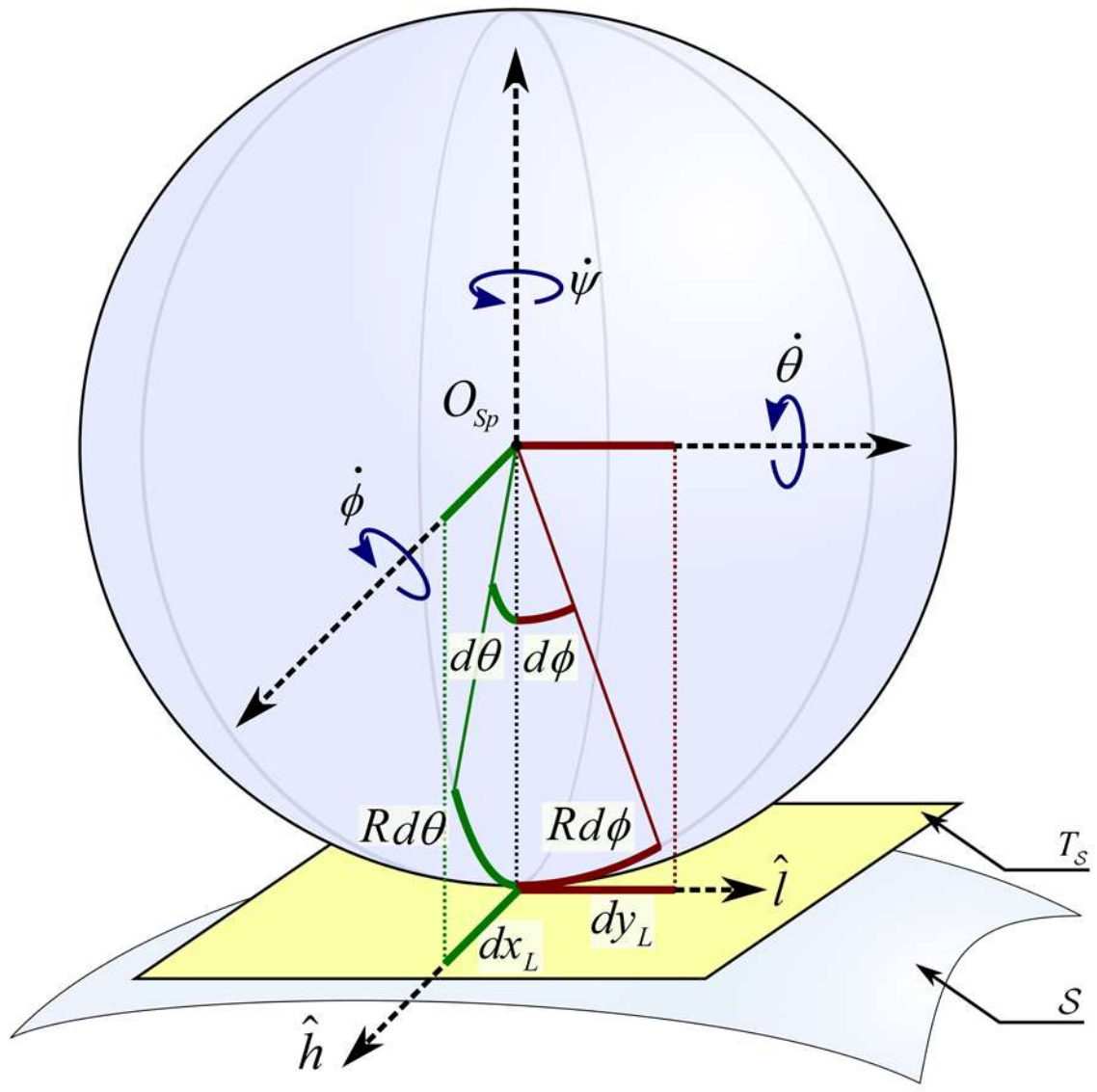

Fig. 4. Differential rolling length of a spherical robot using triple axis rolling during differential of time, $dt$ ,along $\hat{h}$ and $\hat{l}$ directions.

Now, from (28) , (31) and (33) we can calculate elements of ${}^{\mathcal{F}_W}\vec{V}$ as follows:

$$\dot{x} = a_{11}R\dot{\theta} - a_{12}R\dot{\phi}\,,\ \dot{y} = a_{21}R\dot{\theta} - a_{22}R\dot{\phi}\,, \text{ and } \dot{z} = a_{31}R\dot{\theta} - a_{32}R\dot{\phi}\,. \tag{34}$$

Substituting terms presented in (27) in (34), and considering $\dot{\psi}$ as the angular velocity of $\{\mathcal{F}_L\}$ in accordance to $\{\mathcal{F}_T\}$, elements of ${}^{\mathcal{F}_W}\vec{V}$ for a 3R-SR are calculated as the following:

$$\begin{aligned}
\dot{x} &= R\Big(\big(1 - s^2 f_x^{\,2}(1-s_n)\big)C\psi + s^2 f_x f_y (1-s_n) S\psi\Big)\dot{\theta} \\
&\quad - R\Big(\big(1 - s^2 f_x^{\,2}(1-s_n)\big)S\psi - s^2 f_x f_y (1-s_n) C\psi\Big)\dot{\phi}\,, \\
\dot{y} &= -R\Big(s^2 f_x f_y (1-s_n) C\psi + \big(1 - s^2 f_y^{\,2}(1-s_n)\big)S\psi\Big)\dot{\theta} \\
&\quad + R\Big(s^2 f_x f_y (1-s_n) S\psi - \big(1 - s^2 f_y^{\,2}(1-s_n)\big)C\psi\Big)\dot{\phi}\,, \\
\dot{z} &= s_n R\big(f_x C\psi - f_y S\psi\big)\dot{\theta} - s_n R\big(f_x S\psi + f_y C\psi\big)\dot{\phi}\,.
\end{aligned} \tag{35}$$

2) *Dual Axis Rolling Spherical Robot (2R-SR).*

In this subcategory of SRs, the robot provides torques and consequently the ability to roll about its two main axes which are parallel to $T_S$. There are a number of SR designs that relate to this type [23, 32-37] such as hamster wheel SRs. Comparing to 3R-SRs, similar steps should be taken in order to derive kinematics equation of 2R-SRs. The only distinction is that, in 2R-SRs, the robot is not capable of rotating about its vertical axis, i.e., it is assumed that $\dot{\psi}=0$, therefore we can write:

$$\dot{x}=a_{11}R\dot{\theta}-a_{12}R\dot{\phi}\,,\ \dot{y}=a_{21}R\dot{\theta}-a_{22}R\dot{\phi}\,,\ \dot{z}=a_{31}R\dot{\theta}-a_{32}R\dot{\phi}\,,\text{ and }\dot{\psi}=0\,. \tag{36}$$

The term $\dot{\psi}=0$, renders 2R-SRs as robots with a completely different behavior. Assuming $\psi=\psi(0)=0$, for 2R-SRs we can write:

$$\begin{aligned}
&\dot{x}=R\left(1-s^2f_x^2\left(1-s_n\right)\right)\dot{\theta}+R\left(s^2f_xf_y\left(1-s_n\right)\right)\dot{\phi}\,,\\
&\dot{y}=-Rs^2f_xf_y\left(1-s_n\right)\dot{\theta}-R\left(1-s^2f_y^2\left(1-s_n\right)\right)\dot{\phi}\,,\\
&\dot{z}=s_nRf_x\dot{\theta}-s_nRf_y\dot{\phi}\,.
\end{aligned} \tag{37}$$

3) *Rolling and Turning Spherical Robot (RT-SR).*

In addition to rolling action, RT-SRs are able to turn about their vertical axis in order to change their moving direction. In other words, internal parts can provide angular velocities of $\dot{\theta}$ and $\dot{\psi}$, while it is assumed that $\dot{\phi}=0$. RT-SRs are relatively less common but still some designs can be found in the literature [38-40]. To calculate kinematics of RT-SRs one can write:

$$dx_L=Rd\theta\text{, and }dy_L=0\,.\ \dot{x}=a_{11}R\dot{\theta}\,,\ \dot{y}=a_{21}R\dot{\theta}\,,\dot{z}=a_{31}R\dot{\theta}\,. \tag{38}$$

Angular velocity of $\{\mathcal{F}_L\}$ in accordance to $\{\mathcal{F}_T\}$ equals to $\dot{\psi}$. Substituting terms from (27) in (38) we have:

$$\begin{aligned}
&\dot{x}=R\left(\left(1-s^2f_x^2\left(1-s_n\right)\right)C\psi+s^2f_xf_y\left(1-s_n\right)S\psi\right)\dot{\theta}\,,\\
&\dot{y}=-R\left(s^2f_xf_y\left(1-s_n\right)C\psi+\left(1-s^2f_y^2\left(1-s_n\right)\right)S\psi\right)\dot{\theta}\,,\\
&\dot{z}=s_nR\left(f_xC\psi-f_yS\psi\right)\dot{\theta}\,.
\end{aligned} \tag{39}$$

*B. Rolling and Steering Spherical Robots (RS-SR):*

As it is shown in Fig. 5, consider $\dot{\alpha}$ as angular velocity of a RS-SR around its transverse axis that provides forward and backward rolling motion of the robot. In addition, the robot can tilt about its longitudinal axis as its steering mechanism providing turning ability for the robot through an arc shape rolling motion. Unlike CR-SRs, in RS-SRs tilting angle $\phi$ is limited. Another difference is that in RS-SRs the transverse axis is defined in a way that tilts with the robot, so it is not always parallel to the tangent plane. This type of SRs uses an underactuated mechanical propulsion mechanism that allows the robot to roll in a nonholonomic manner [41-44].

Fig. 5 illustrates a RS-SR with tilting angle of $\phi$ with respect to the tangent plane. The turning action of the sphere can be modeled as a cone with apex angle of $2\phi$, purely rolling over the surface $\mathcal{S}$, for which the instantaneous axis of rotation is the imaginary line passing through cone's vertex $V$, and $P_0$ that lies in the tangent plane $T_S$. Considering the instantaneous differential rolling arc of the cone's base circle, we have:

$$r_c d\alpha = \rho d\psi \,, \tag{40}$$

where, $r_c$ is the cone's base circle radius, $d\alpha$ and $d\phi$ are differential rotation of the sphere during differential of time, $dt$, about the transverse and longitudinal axis respectively. $\rho$ is the instantaneous radius of curvature of the turning motion that is the distance between the vertex of the cone and the contact point. These two quantities can be calculated as:

$$r_c = R\cos(\phi) \,, \tag{41}$$

and,

$$\rho = -R\cot(\phi) \,. \tag{42}$$

Substituting (41) and (42) into (40), results in:

$$d\psi = -d\alpha \sin(\phi) \,. \tag{43}$$

Dividing both sides of (43) by $dt$, we have:

$$\dot{\psi} = -\dot{\alpha}\sin(\phi) \,. \tag{44}$$

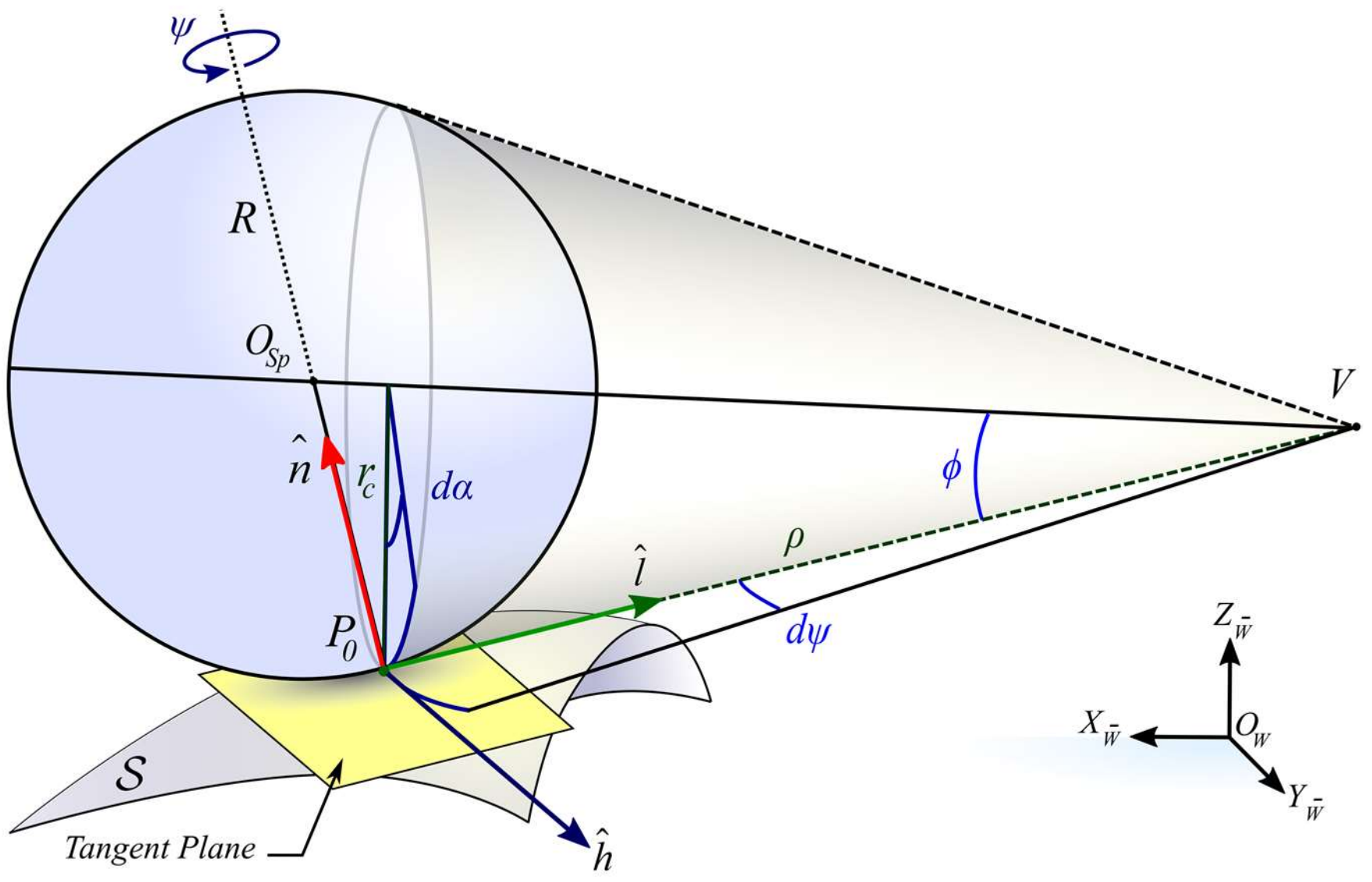

Figure 5. Schematic diagram of a spherical robot using rolling and steering (RS) action kinematics.

Moving on, the position vector of the center point of the sphere in $\{\mathcal{F}_L\}$ in accordance with $P_0$ can be written as:

$${}^{\mathcal{F}_L}\vec{r}_{P_0 O_{Sp}} = R\hat{n} \,, \tag{45}$$

and angular velocity of the sphere in $\{\mathcal{F}_L\}$ can be written as:

$$ {}^{\mathcal{F}_L}\vec{\Omega}_{Sp} = \dot{\phi}\hat{h} + \dot{\alpha}\cos(\phi)\hat{l} - \dot{\alpha}\sin(\phi)\hat{n} . \quad (46) $$

Then, the linear velocity of the sphere's center point can be calculated as follows:

$$ {}^{\mathcal{F}_L}\vec{V}_{Sp} = {}^{\mathcal{F}_L}\vec{\Omega}_{Sp} \times {}^{\mathcal{F}_L}\vec{r}_{P_0 O_{Sp}} = R\dot{\alpha}\cos(\phi)\hat{h} - R\dot{\phi}\hat{l} . \quad (47) $$

Forthwith, we are able to form ${}^{\mathcal{F}_L}\vec{V}$ as the following:

$$ {}^{\mathcal{F}_L}\vec{V} = [R\dot{\alpha}\cos(\phi), -R\dot{\phi}, 0]^T . \quad (48) $$

Eventually, from (31) and (48) we can calculate components of ${}^{\mathcal{F}_W}\vec{V}$ as follows:

$$ \begin{aligned} &\dot{x} = a_{11}R\dot{\alpha}\cos(\phi) - a_{12}R\dot{\phi} , \;\; \dot{y} = a_{21}R\dot{\alpha}\cos(\phi) - a_{22}R\dot{\phi} \\ &\dot{z} = a_{31}R\dot{\alpha}\cos(\phi) - a_{32}R\dot{\phi} . \end{aligned} \quad (49) $$

Equations in (49), using terms presented in (27), can be utilized to calculate the position of the SR, $P_0$, based on input parameters of robot, $\dot{\alpha}$ and $\phi$ as the following:

$$ \begin{aligned} &\dot{x} = R\Big(\big(1 - s^2 f_x^{\,2}(1-s_n)\big)C\psi + s^2 f_x f_y (1-s_n) S\psi\Big)\dot{\alpha}C\phi \; - R\Big(\big(1 - s^2 f_x^{\,2}(1-s_n)\big)S\psi - s^2 f_x f_y (1-s_n) C\psi\Big)\dot{\phi} , \\ &\dot{y} = -R\Big(s^2 f_x f_y (1-s_n) C\psi + \big(1 - s^2 f_y^{\,2}(1-s_n)\big)S\psi\Big)\dot{\alpha}C\phi \; + R\Big(s^2 f_x f_y (1-s_n) S\psi - \big(1 - s^2 f_y^{\,2}(1-s_n)\big)C\psi\Big)\dot{\phi} \\ &\dot{z} = s_n R\big(f_x C\psi - f_y S\psi\big)\dot{\alpha}C\phi - s_n R\big(f_x S\psi + f_y C\psi\big)\dot{\phi} . \end{aligned} \quad (50) $$

In all types of SR, to calculate the position of the center point of the sphere denoted as $O_{Sp} = [X_{Sp}, Y_{Sp}, Z_{Sp}]^T$ we have:

$$ O_{Sp} = P_0 + R\hat{n} . \quad (51) $$

Alternatively, utilizing (26) and using ${}^{\mathcal{F}_L}O_{Sp} = [0,0,R,1]^T$, it can be calculated as:

$$ O_{Sp} = \mathcal{T}_{LW}\, {}^{\mathcal{F}_L}O_{Sp} = \begin{bmatrix} a_{13}R + a_{14} & a_{23}R + a_{24} & a_{33}R + a_{34} & 1 \end{bmatrix} . \quad (52) $$

Observing terms in (27), one can notice that $P_0 = [a_{14}, a_{24}, a_{34}]^T$ and $\hat{n} = [a_{13}, a_{23}, a_{33}]^T$, therefore, (52) is analogous to (51).

## 5. Control of Spherical Robots over 3D Surfaces

In this section, the pure pursuit method is utilized to address the path tracking problem of the four different types of spherical robots on a 3D surface. Considering $\mathcal{P} : [x_d(t), y_d(t), z_d(t)]^T$ as a 3D desired trajectory lying on $\mathcal{S}$, that is assumed to be smooth, i.e. $\| [\dot{x}_d, \dot{y}_d, \dot{z}_d]^T \|$ is bounded for $t > 0$. It is desired that the location of $P_0 = [x_0(t), y_0(t), z_0(t)]^T$, converges to $\mathcal{P}$ with a stable error. $e$ is the tracking error defined as:

$$ e = [x_d(t), y_d(t), z_d(t)]^T - P_0 . \quad (53) $$

Additionally, defining deviation angle $\zeta$ as the angle between the heading direction, $\hat{h}$, and $e$, we have:

$$ \zeta = \angle(e, \hat{h}) . \quad (54) $$

The objective of this algorithm is to converge the tracking error $e$ along with $\zeta$ to zero for all four proposed categories of SRs. To that end, in each category the path planning algorithm computes values for actuated kinematics states, namely $\dot{\theta}$, $\dot{\phi}$, $\dot{\alpha}$, and $\dot{\psi}$.

*A. 3R-SR Kinematics Control*

As 3R-SRs are capable of providing $\dot{\theta}$, $\dot{\phi}$, and $\dot{\psi}$, the controller is designed to provide required values accordingly as follows:

$$\begin{aligned}&\dot{\theta}=k_{\theta}\frac{\|e\|}{(k_e+\|e\|)}\cos(\zeta)+\frac{1}{R}\|[\dot{x}_d,\dot{y}_d,\dot{z}_d]^T\|,\\&\dot{\phi}=-\left(k_{\phi_1}\frac{\|e\|}{(k_e+\|e\|)}\sin(\zeta)+k_{\phi_2}\sin(\zeta)\right),\text{ and }\dot{\psi}=k_{\psi}\zeta.\end{aligned} \tag{55}$$

In the design of $\dot{\theta}$, $\|e\|/(k_e+\|e\|)$ is used to normalize the error gain to provide smooth transition for control efforts from longer distances to the vicinity of the target point, thus, to compensate the magnitude of $e$, the first term of $\dot{\theta}$ uses $\cos(\zeta)$ multiplied with a normalized error gain. $\cos(\zeta)$ is positive or negative when the SR is behind or ahead of the target respectively. Additionally, one can easily recognize that $\|[\dot{x}_d,\dot{y}_d,\dot{z}_d]^T\|/R$ is the angular velocity that is required for the sphere to pursue the target based on its speed. Therefore, this term is used as the bias value for $\dot{\theta}$. For $\dot{\phi}$, if $\zeta \gg 0$, i.e., the robot is not heading towards the target point, the robot tries to compensate its lateral distance to the target by rolling laterally using $\dot{\phi}$ in parallel to other actuators. If the robot is heading towards the target, $\zeta \approx 0$, the robot relies mostly on $\dot{\theta}$ to approach the target. Finally, $\dot{\psi}$ is designed to be directly proportional to $\zeta$ as a measure of required turning to compensate the deviation angle.

*B. 2R-SR Kinematics Control*

Considering the fact that 2R-SRs are not able to turn, i.e. $\dot{\psi}=0$, it can be implied that a 2R-SR is not capable of following the target by changing its heading direction and consequently the robot can only use $\dot{\theta}$ and $\dot{\phi}$ to move towards the target. Therefore, the controller is designed such that:

$$\begin{aligned}&\dot{\theta}=k_{\theta_1}\frac{\|e\|}{(k_e+\|e\|)}\cos(\zeta)+k_{\theta_2}\cos(\zeta),\\&\dot{\phi}=-\left(k_{\phi_1}\frac{\|e\|}{(k_e+\|e\|)}\sin(\zeta)+k_{\phi_2}\sin(\zeta)\right),\ \dot{\psi}=0.\end{aligned} \tag{56}$$

### C. RT-SR Kinematics Control

RT-SRs can move towards the target point by compensating their deviation error, $\zeta$ through turning action, and then using $\dot{\theta}$ to approach and follow the target. This method is used to design the controller to provide required angular velocities as the following:

$$\dot{\theta}=k_{\theta}\frac{\|e\|}{(k_e+\|e\|)}\cos(\zeta)+\frac{1}{R}\|[\dot{x}_d,\dot{y}_d,\dot{z}_d]^T\|,\ \dot{\phi}=0,\ \dot{\psi}=k_{\psi}\zeta. \tag{57}$$

### D. RS-SR Kinematics Control

Calculated values for $\dot{\alpha}$ and $\dot{\phi}$ are as the following for RS-SRs:

$$\dot{\alpha}=k_{\alpha}\frac{\|e\|}{(k_e+\|e\|)}\cos(\zeta)+\frac{1}{R}\|[\dot{x}_d,\dot{y}_d,\dot{z}_d]^T\|,\ \dot{\phi}=k_{\phi}\zeta. \tag{58}$$

It should be noted that $\dot{\psi}$ is driven and cannot be used as actuated value in RS-SRs.

## 6. SIMULATION RESULT

This section provides the simulation results through MATLAB/Simulink to verify the presented method and the controllers that are proposed in section 5. The simulation is carried out, where for the terrain surface, $\mathcal{S}$ is defined as the following:

$$z=f(x,y)=a(\cos(\omega x)+\cos(\omega y)-2), \tag{59}$$

where,

$$x(t)=2\cos(5t/100) \text{ and } y(t)=2\cos(5t/100) \tag{60}$$

Additionally, the following numerical parameters are used. $a=0.2m$, $\omega=2$, and $R=0.2m$ and controller design parameters are tuned by trial-and-error.

It can be observed from Figs 6 and 7 that the SRs can track the desired path successfully, and the tracking error converges to the vicinity of the origin in approximately 30 seconds. Fig. 6 shows a 3D representation of all types of SRs (CR (3R, 2R, RT), and RS), rolling over the surface defined in (59).

Fig. 7 represents the actual and desired values for $X$, $Y$, and $Z$ against time for the robots. It can be seen that the tracking error is bounded due to the fact that $\|[\dot{x}_d,\dot{y}_d,\dot{z}_d]^T\|$ is bounded. Since the desired trajectory is unknown to the pure pursuit, the tracking error remains bounded, while the convergence bound can reduced by tuning to any desired level. According to the simulation results presented in this section, all types of SRs can effectively track the desired 3D trajectory by utilizing the designed controllers.

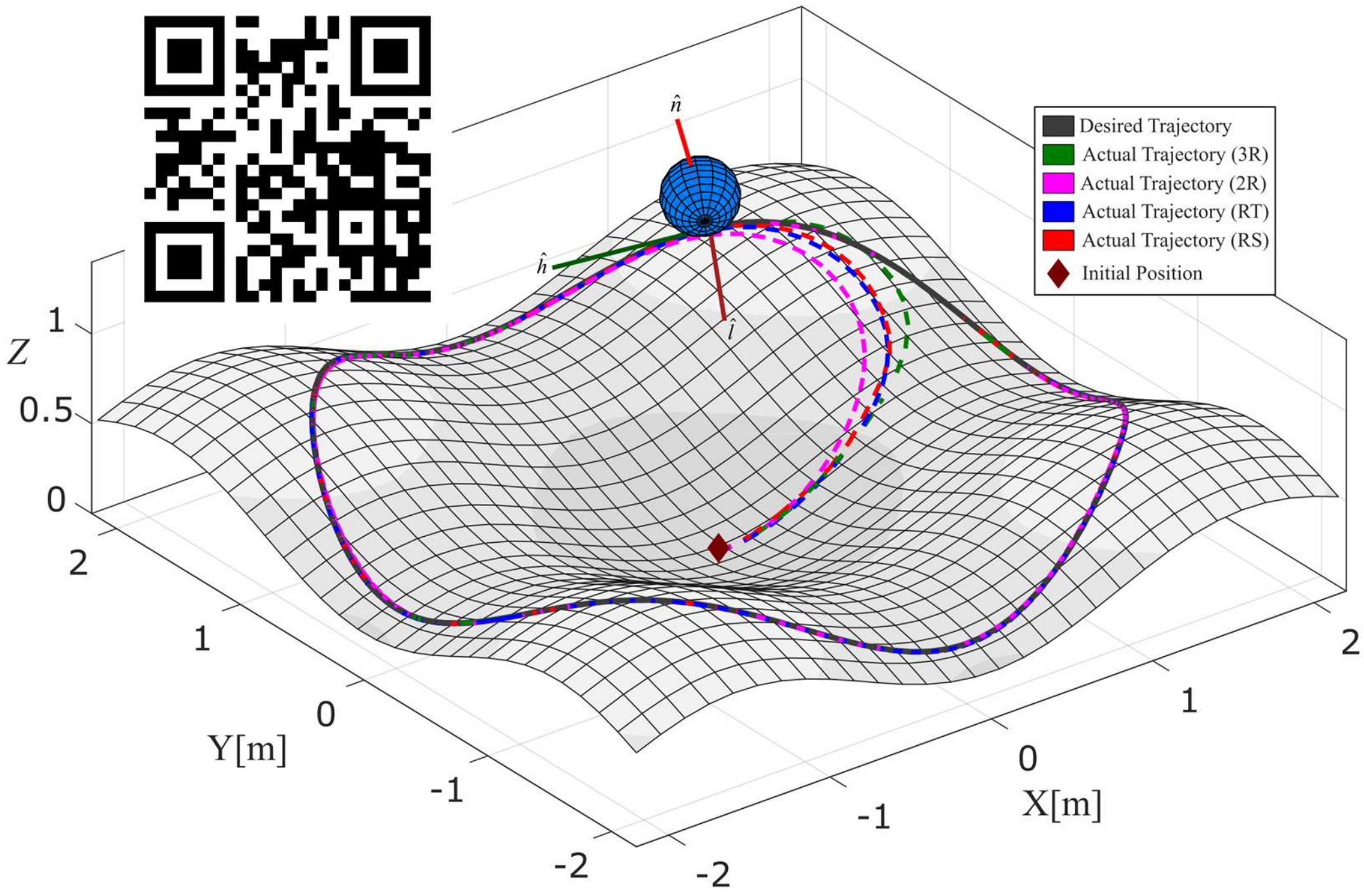

Figure 6. Robots' actual and desired trajectory on a 3D surface. Scan the QR-code to watch an animation of the same simulation result.

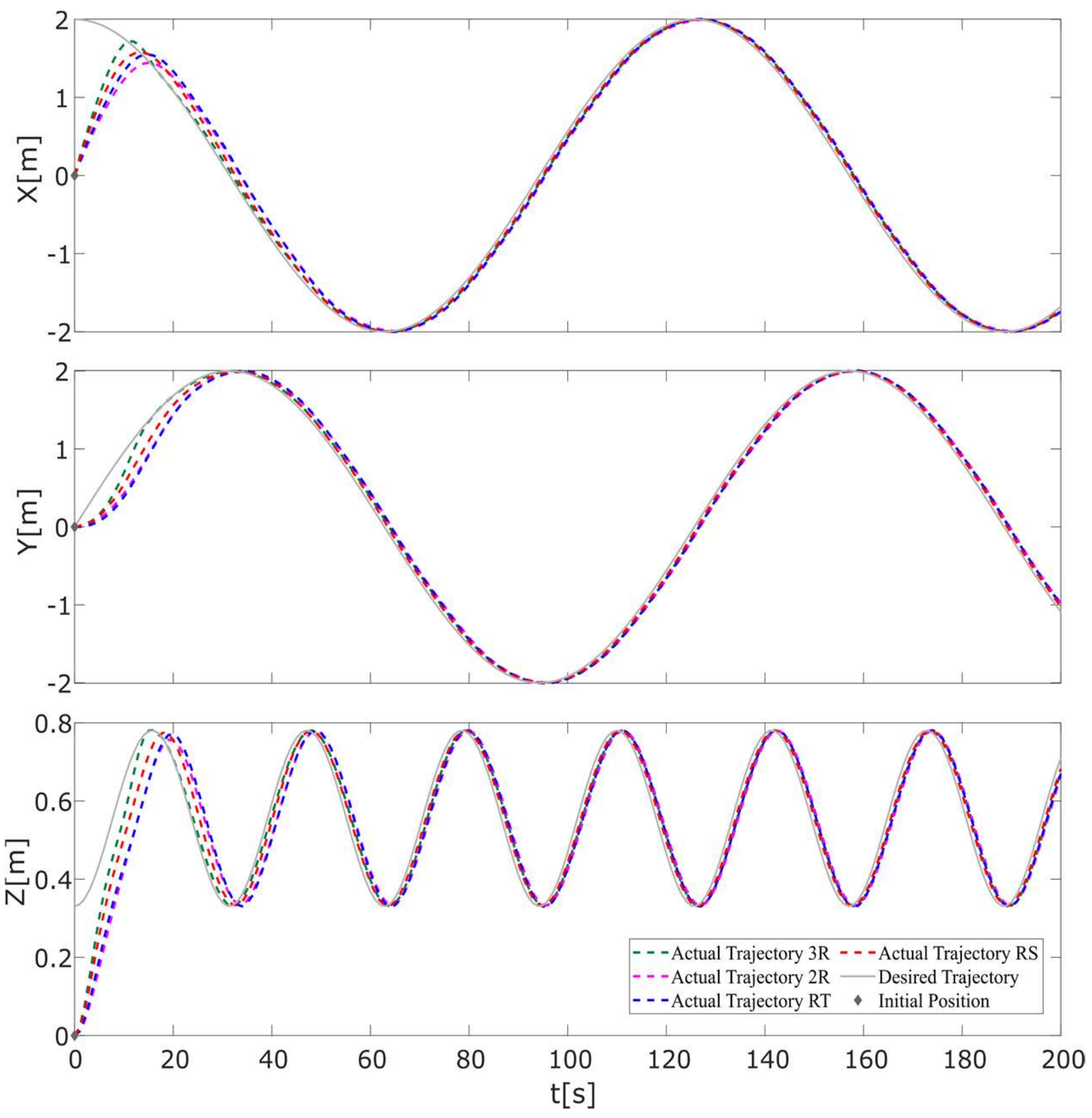

Figure 7. Robots' actual and desired trajectory along $X$, $Y$, and $Z$ directions versus time.

## 7. Conclusion

This paper addressed the problem of modeling the kinematics and path tracking control of popular SR configurations rolling over 3D terrains. It is noteworthy that the aim of this paper is not to compare the performance of different types of spherical robots. In fact, it provides a classification of SRs based on their kinematics behavior to derive kinematics equations on 3D surfaces. Utilizing the proposed kinematics, one can use the Euler-Lagrange method to derive the dynamics of robots on uneven terrains by keeping in mind that the gravity direction is not always perpendicular to the tangent plane.

## Appendix A

In this appendix, calculation steps of utilizing rotation quaternions instead of Rodriguez rotation method to calculate $\mathcal{R}_{TTr}$ are presented. Generally, if we write unit rotation axis vector $\hat{e}$ and $\hat{i}_{Tr}$ in the form of:

$$q(\hat{e})=\cos(\gamma/2)+\left(e_x i+e_y j+0k\right)\sin(\gamma/2), \tag{A1}$$

and $q(\hat{i}_{Tr})=[0,1,0,0]^T$, then it can be shown that $\hat{i}_T$ is calculated as an operation called conjugation of $q(\hat{i}_{Tr})$ by $q(\hat{e})$ defined as follows:

$$\hat{i}_T=q(\hat{e})q(\hat{i}_{Tr})q(\hat{e})^{-1}. \tag{A2}$$

we know that as $\|q(\hat{e})\|=1$, then:

$$q(\hat{e})^{-1}=\cos(\gamma/2)-\left(e_x i+e_y j\right)\sin(\gamma/2). \tag{A3}$$

Operation (A2) can be performed through Hamilton product in which $q(\hat{e})$, $q(\hat{i}_{Tr})$, and $q(\hat{e})^{-1}$ are written in $1\times 4$ quaternions format and the distributive law is used over all elements. However, a more convenient method can be utilized in matrix format. Any unit quaternion rotation in the form of $p'=qpq^{-1}$ with $q=(q_r,q_v)$, $q_v=q_i i+q_j j+q_k k$ can be written in the form of rotation matrix $p'=\mathcal{R}p$ as the following:

$$\mathcal{R}=q_v\otimes q_v+q_r{}^2\mathbf{I}+2q_r E_\times+E_\times{}^2, \tag{A4}$$

where, $\otimes$ denotes outer product operation. Equation (A4) can be written in the expanded format as the following:

$$\mathcal{R}_{TTr}=\begin{bmatrix}1-2(q_j^2+q_k^2) & 2(q_iq_j-q_kq_r) & 2(q_iq_k-q_jq_r)\\ 2(q_iq_j-q_kq_r) & 1-2(q_i^2+q_k^2) & 2(q_jq_k-q_iq_r)\\ 2(q_iq_k-q_jq_r) & 2(q_jq_k-q_iq_r) & 1-2(q_i^2+q_j^2)\end{bmatrix}. \tag{A5}$$

Substituting $q_k=0$ in (A5) results in:

$$\mathcal{R}_{TTr}=\begin{bmatrix}1-2q_j^2 & 2q_iq_j & -2q_jq_r\\ 2q_iq_j & 1-2q_i^2 & -2q_iq_r\\ -2q_jq_r & -2q_iq_r & 1-2(q_i^2+q_j^2)\end{bmatrix}. \tag{A6}$$

While, $q_r=\cos(\gamma/2)$, $q_i=sf_y\sin(\gamma/2)$, and $q_j=-sf_x\sin(\gamma/2)$, it can be shown that, plugging these terms in (A6) results in (20).